# Xtreme Margin: A Tunable Loss Function for Binary Classification Problems

**Rayan S. Wali**[1,2]

[1] Cornell University
[2] rsw244@cornell.edu

Loss functions drive the optimization of machine learning algorithms. The choice of a loss function can have a significant impact on the training of a model, and how the model learns the data.

Binary classification is one of the major pillars of machine learning problems. It is used in medical imaging to failure detection applications. The most commonly used surrogate loss functions for binary classification include the binary cross-entropy and the hinge loss functions, which form the focus of our study.

In this paper, we provide an overview of a novel loss function, the Xtreme Margin loss function. Unlike the binary cross-entropy loss, this loss function is tunable with hyperparameters $\lambda_1$, $\lambda_2$, i.e., changing their values will alter the training of a model.

## 1. INTRODUCTION

The binary cross-entropy loss function is a function applied to the predicted probability score of a binary classification model and the true label for a particular instance. It learns the probability distribution $\mathcal{P}(y_i|\vec{x})$, for a feature vector $\vec{x}$ and the predicted label of the $i$th class, $i \in \{0, 1\}$, $y_i$. It can be mathematically expressed as follows for an instance $i$:

$$\mathcal{L} = -\left(y_i \cdot log(p(x_i)) + (1 - y_i) \cdot log(1 - p(x_i))\right). \quad \textbf{(1)}$$

The probability score $p(x_i)$ is the predicted probability that instance $x_i$ belongs to the default class and $y_i$ is the true label, either 0 or 1, for instance $i$.

A similar loss function, the hinge loss function, is also used for binary classification problems, particularly for training Support Vector Machine (SVM) models. The hinge loss function attempts to maximize the *margin*, whereas the objective of binary cross-entropy is to maximize a likelihood function. It is also more computationally efficient during training compared to the binary cross-entropy loss.

In this paper, we introduce a third loss function for binary classification models, *Xtreme Margin*, that adds a tunable component that allows practitioners to control the *willingness* of a data point to be part of the default or the non-default class.

## 2. RELATED WORK

Regularization techniques are studied to allow for increased flexibility during training by minimizing the model coefficients, ultimately reducing overfitting. However, regularization, such as L1 or L2 regularization, is applied on a *cost function*, whereas this paper examines adding a tunable hyperparameter component to a *loss function*. The other notable difference is that the regularizer $\lambda$ in regularization is placed to minimize the model parameters as follows:

$$min_\theta \left[ \sum_{i=1}^{N} \mathcal{L}(\theta) + \lambda \sum_{i=1}^{M} (\theta_i)^2 \right]. \quad \textbf{(2)}$$

However, the proposed method in this paper is to add two regularizer terms to control the *margin*, which in this context, is defined as the difference in confidence scores between the two classes, as well as to control the willingness of a data point to be part of a particular class. This particular manipulation introduced in this paper has not been previously studied.

The bias-variance tradeoff states that as the bias decreases, the variance increases. The bias can be measured by the accuracy, whereas the variance can be measured by the intra-class variability. Through Xtreme Margin, researchers and practitioners will have the control to adjust the bias-variance tradeoff by changing the hyperparameters $\lambda_1$ and $\lambda_2$.

Regularizing a machine learning model is studied to increase the bias of the model. Similarly, reducing the number of layers of a neural network to prevent overfitting increases the bias of the network. We will study methods to mitigate this problem with our proposed tunable Xtreme Margin loss function. This allows the tradeoff between validation accuracy and the model bias to be controllable with this parameter. It is also studied that a problem with a low number of features used during model training can increase the bias of the model. For the sake of our experiments, we will use a small set of features for training our models to demonstrate that the bias can be improved, even in this situation, through the loss function.

## 3. THE XTREME MARGIN LOSS FUNCTION

The Xtreme Margin loss function is defined as follows:

$$\mathcal{L}(y, y_{true}; \lambda_1, \lambda_2) = \frac{1}{1 + (\sigma(y, y_{true}) + \gamma)}. \quad (3)$$

Below, we expand the $\gamma$ term above. The $\lambda(2y - 1)^2$ term of the expression below is the extreme margin term, and is derived from the squared difference between the true conditional probability prediction score of belonging to the default class and the true conditional probability prediction score of belonging to the non-default class. For example, if $y = 0.80$ and the default class is '1', then the extreme margin term will be $(0.80 - (1 - 0.80))^2 = 0.36$.

$$\gamma = \mathbb{1}_{[ytrue=ypred \& ytrue=0]} \times \lambda_1 (2y - 1)^2 \quad (4)$$

$$+ \mathbb{1}_{[ytrue=ypred \& ytrue=1]} \times \lambda_2 (2y - 1)^2. $$

$$\mathbb{1}_A(x) := \begin{cases} 1 & \text{if } x \in A \,, \\ 0 & \text{if } x \notin A \end{cases}, \quad (5)$$

$$\sigma(y, y_{true}) := \begin{cases} 0 & \text{if } |y - y_{true}| < 0.5, \\ \frac{1}{e|y_{true} - y|} - 1 & \text{if } |y - y_{true}| \geq 0.5 \end{cases}, \quad (6)$$

$$y_{pred} := \begin{cases} 1 & \text{if } y \geq 0.50 \,, \\ 0 & \text{if } y < 0.50 \end{cases}. \quad (7)$$

## 4. ANALYSIS

As discussed, loss functions are the metrics that are used to train machine learning models. Even though most times we look at accuracy as the standard metric, it is actually not used to train the model. Hence, it is critical to understand all the components of the loss function in order to have a complete understanding of how it scores an output given the label for an instance.

### A. Simple Evaluation

The 'simple evaluation' is performed by the $\sigma(y, y_{true})$ component of the equation above. This expression is 0 when the predicted value is correct, and is $\frac{1}{e|y_{true} - y|} - 1$ when the predicted value is incorrect. *(i)* When the predicted value is correct, the loss is taken care solely by the extreme evaluation which we will discuss below. *(ii)* When the predicted value in incorrect, the loss score is higher. As the gap between $y$ and the true value $y_{true}$ increases, the misclassification is worse, and the term $|y_{true} - y|$ increases, decreasing the entire $\sigma(y, y_{true})$ expression. Since this is all part of the denominator, the overall loss score increases as expected.

### B. Extreme Evaluation

The 'extreme evaluation' is performed by the $\lambda(2y - 1)^2$ component of the equation above. This component measures the difference between the outputted probability for the default class and the (indirectly) outputted probability for the non-default class. For instance, if the model predicts a 80% chance that an instance belongs to class '1', the component $2y - 1$ represents the difference between 80% and the 20% probability of the instance belonging to class '0'.

### C. Range of Loss Values

The range of values that $\mathcal{L}$ can output is $(0, e]$. To show this, we consider the following cases:

---

**Case 1.** *The true value for an instance is not equal to its predicted value.*

**Analysis.** The indicator function will be 0, and the denominator will be only left with the smoothing factor of 1 added to the result of the $\sigma$ function. When the true value of an instance is not equal to its predicted value, $\sigma(y, y_{true}) = \frac{1}{e|y_{true} - y|} - 1$.

$$\mathcal{L} = \frac{1}{1 + \frac{1}{e|y_{true} - y|} - 1} = e|y_{true} - y|.$$

Since $|y_{true} - y|$ can be 1 at maximum, the maximum value $\mathcal{L}$ can be is $e$.

---

**Case 2.** *The true value for an instance is equal to its predicted value.*

**Analysis.** In this case, the term multiplied with the indicator function does not zero-out. The $\sigma$ term takes the model's level of confidence into account, and is fixed regardless of the value of the hyperparameter $\lambda$. The term multiplied with $\lambda$ represents the (only) tunable component of the loss function — increasing the value of $\lambda$ will place more emphasis on reducing the gap in the output conditional probabilities for both classes. The minimum $\mathcal{L}$ can be is 0, since the following limit holds:

$$\lim_{\lambda \to \infty} \mathcal{L}(\lambda) = 0.$$

---

In other words, the term with the hyperparameter $\lambda$ minimizes the following expression:

$$|\mathcal{P}(Y = 1 \mid X = \vec{x}) - \mathcal{P}(Y = 0 \mid X = \vec{x})|. \quad (8)$$

Essentially, $\lambda$ controls the tradeoff between the plain model confidence and extreme model confidence. Hence, when this loss function is used, it is good practice to use hyperparameter tuning to find the optimal parameter $\lambda$ for your problem to control the variance in the performance across the different instances.

## 5. TENSORFLOW IMPLEMENTATION

The following code is the implementation of the Xtreme Margin loss function in TensorFlow for a batch size of 1:

```python
import tensorflow as tf
from keras import backend as K

lmbda1 = ### User-defined hyperparameter
lmbda2 = ### User-defined hyperparameter

def indicator1(y_true, y_pred):
    if tf.equal(tf.dtypes.cast(y_true, tf.float32),
        y_pred) and tf.equal(tf.dtypes.cast(y_true,
        tf.float32), tf.constant(0.)):
        return tf.constant(1.)
    else: return tf.constant(0.)

def indicator2(y_true, y_pred):
    if tf.equal(tf.dtypes.cast(y_true, tf.float32),
        y_pred) and tf.equal(tf.dtypes.cast(y_true,
        tf.float32), tf.constant(1.)):
        return tf.constant(1.)
    else: return tf.constant(0.)

def sigma(y, y_true):
    if tf.less(tf.abs(tf.subtract(y, tf.dtypes.cast(
        y_true, tf.float32))), 0.5):
        out = 0.
    else:
        out = tf.subtract(tf.divide(1, tf.multiply(
            2.718, tf.abs(tf.subtract(tf.dtypes.
            cast(y_true, tf.float32), y)))), 1.)
    return out

def xtreme_margin_loss(y_true, y):
    y_pred = tf.reshape(tf.constant(1.), [1,1])
        if tf.equal(tf.greater(y, tf.constant(0.5)),
            True) else tf.reshape(tf.constant(0.),
            [1,1])
    loss = tf.divide(1., tf.add(1., tf.add(sigma(
        y, y_true), tf.add(tf.multiply(indicator1(
        y_true, y_pred), tf.multiply(lmbda1,
        tf.square(tf.subtract(tf.multiply(2., y),
        1.)))), tf.multiply(indicator2(y_true,
        y_pred), tf.multiply(lmbda2, tf.square(
        tf.subtract(tf.multiply(2., y), 1.)))))))))
    return K.mean(loss, axis=-1)
```

## 6. INSTANCE PENALIZATION

If an instance is correctly predicted, it should ideally be assigned a low loss score, and if it is incorrectly predicted, it should be assigned a high loss score. Binary cross-entropy penalizes instances based on their predicted probability in comparison to their true class. The hinge loss function penalizes instances based on their misclassifications as well as their margin to the decision boundary representing the trained model. The Xtreme Margin loss function penalizes instances the following way:

- **Case 1.** (an instance is correctly predicted and belongs to non-default class) The loss score is defined by the extreme margin term $\lambda_1(2y - 1)^2$. It is composed of a user-defined value and the classification model's prediction.

- **Case 2.** (an instance is correctly predicted and belongs to default class) The loss score is defined by the extreme margin term $\lambda_2(2y - 1)^2$. It is composed of a user-defined value and the classification model's prediction.

- **Case 3.** (an instance is incorrectly predicted) The loss score is $e|y_{true} - y|$, which is higher than the loss scores in the first two cases.

The hyperparameters $\lambda_1$ and $\lambda_2$ control how much weight should be assigned to correctly classifying an instance belonging to a particular true class.

- **Case 1.** ($\lambda_1 = \lambda_2$) In this case, the training algorithm will equally score an instance belonging to the default class and an instance belonging to the non-default class assuming that they are equally distant to the true class (e.g., an instance belonging to class '0' has a conditional probability of 0.20 and an instance belonging to class '1' has a conditional probability of 0.80).

- **Case 2.** ($\lambda_1 \ll \lambda_2$) In this case, the training algorithm will strongly bias the default class. The training algorithm will try to correctly predict the training instances belonging to the default class since they yield a low loss score.

- **Case 3.** ($\lambda_1 \gg \lambda_2$) In this case, the training algorithm will strongly bias the non-default class. The training algorithm will try to correctly predict the training instances belonging to the non-default class since they yield a low loss score.

## 7. EXPERIMENT SETUP

In order to evaluate the Xtreme Margin loss function, we compare it to the binary cross-entropy and the hinge losses on two binary classification datasets: (1) the `sonar` dataset and (2) the `ionosphere` dataset. We will measure the mean cross-validation score and the standard deviation of the cross-validation scores for the $k$ stratified splits ($k = 10$) on the Sonar dataset. We will then repeat this cross-validation process for 20 iterations and report the minimum and maximum cross-validation means and standard deviations in Table 1 below. We will then use the Ionosphere dataset to plot the decision boundaries and the training and testing accuracy plots.

The following is the artificial neural network model that is held constant throughout our experimentation below.

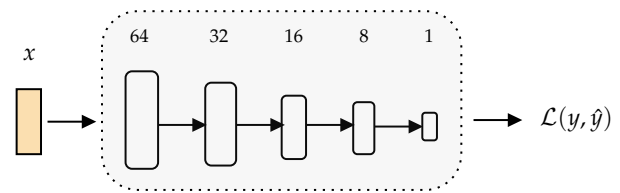

```python
model = Sequential()
model.add(Dense(64, activation='relu'))
model.add(Dropout(0.25))
model.add(Dense(32, activation='sigmoid'))
model.add(Dropout(0.25))
model.add(Dense(16, activation='relu'))
model.add(Dropout(0.25))
model.add(Dense(8, activation='relu'))
model.add(Dense(1, activation='sigmoid'))
```

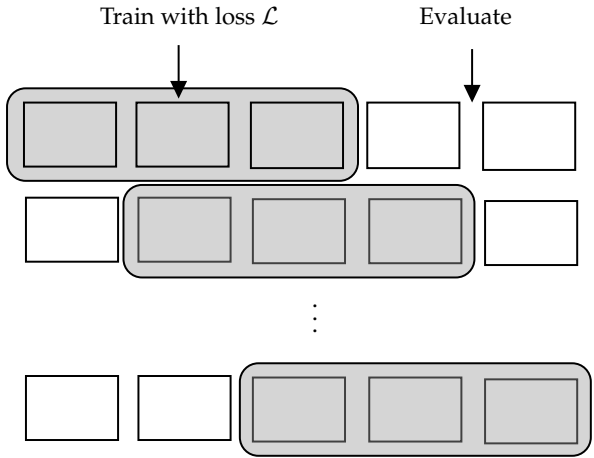

cross-validation scores versus values of lambda to visualize how changing $\lambda$ affects an unbiased estimator of the test error.

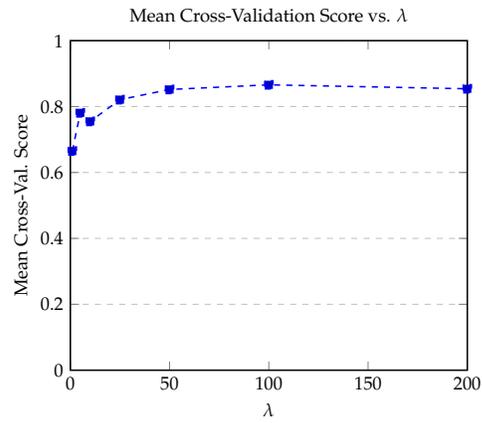

**Fig. 1.** Hyperparameter tuning $\lambda$ on mean cross val. score

## 8. PERFORMANCE COMPARISONS

Table 1 below shows the performances for the different loss functions on the same problem, holding all other factors constant, such as the model (identical neural network architectures), optimizer type (RMSprop), and the number of epochs trained for (100). For the Xtreme Margin loss, we will set $\lambda_1 = \lambda_2$ for the sake of simplicity.

**Table 1.** Performances for Loss Functions ($\lambda_1 = 1, \lambda_2 = 1$)

| Loss Function | Mean Cross-Val. Score | Std. Cross-Val. Score |
|---|---|---|
| Binary Cross-Entropy | $0.8269 - 0.8598$ | $0.0326 - 0.0958$ |
| Hinge | $0.7498 - 0.8179$ | $0.0477 - 0.1643$ |
| Xtreme Margin | $0.7786 - 0.8512$ | $0.0292 - 0.1323$ |

We can see that the Xtreme Margin loss function with default hyperparameter $\lambda = 1$ achieved better lower-bound and upper-bound average cross-validation scores than the hinge loss function but worse than the binary cross-entropy loss function. However, the category that the Xtreme Margin loss function was the comparative winner was in the lower bound of the standard deviation of the cross-validation score. The Xtreme Margin achieved a lower cross-validation standard deviation lower bound relative to all tested models, implying that the losses for the different instances in the dataset did not have as much variability in the best case. In the worst case, it was outperformed by the binary cross-entropy loss function.

Note that the performances above were for the untuned Xtreme Margin loss. After tuning the hyperparameter $\lambda$ with grid search, we obtain the following results:

**Table 2.** Performances for Loss Functions ($\lambda_1^* = 100$, $\lambda_2^* = 100$)

| Loss Function | Mean Cross-Val. Score | Std. Cross-Val. Score |
|---|---|---|
| Binary Cross-Entropy | $0.8269 - 0.8598$ | $0.0326 - 0.0958$ |
| Hinge | $0.7498 - 0.8179$ | $0.0477 - 0.1643$ |
| Xtreme Margin | $0.8062 - 0.8652$ | $0.0348 - 0.0982$ |

Below, we plot the mean and standard deviation

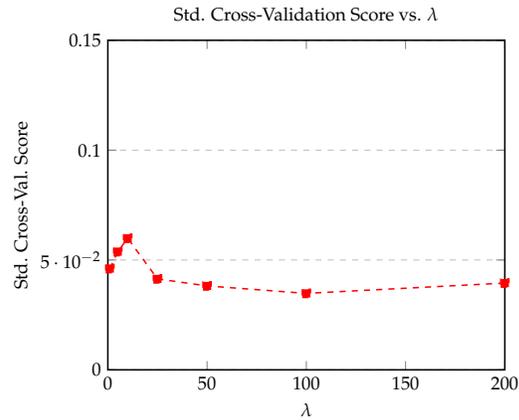

**Fig. 2.** Hyperparameter tuning $\lambda$ on std. cross val. score

For the machine learning problem tested on, the optimal value of $\lambda$ was approximately 100, determined through hyperparameter tuning. We can see this in both graphs above, as the mean cross-validation score is maximized at this value and the standard deviation score is minimized at this value. This technique can be applied to any problem to find the optimal values of $\lambda_1$ and $\lambda_2$ that best train your model.

## 9. CONVEXITY ANALYSIS

Convexity is a property of functions that measures their degree of curvature. They play an important role in the training on loss functions. Convex loss functions are desirable as they bring convergence to the training process when optimized with techniques such as gradient descent. However, gradient descent methods are proven to optimize a loss function well even with non-convex loss functions. Stochastic gradient descent, when applied to a non-convex loss function, is studied to converge. In their paper, Zhao et al. propose a training algorithm for non-convex optimization that can optimize a non-convex loss function, Ramp Loss, and achieve a reasonably high training accuracy and a test accuracy higher than the other convex and non-convex loss functions compared in the study [32].

We will determine the convexity of our loss function $\mathcal{L}$ decomposed on the prediction. There are several ways to measure the convexity of a function, but the one we will consider below is through directly plotting a reduced version of the functions. Consider the following two cases:

**Case 1.** $y_{true} = y_{pred}$ $(|y - y_{true}| < 0.5)$

$$\mathcal{L} = \frac{1}{1 + (2y - 1)^2}.$$

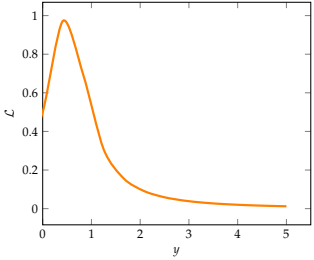

**Fig. 3.** Loss Function $\mathcal{L}$ vs. $y$ — Case 1

**Case 2.** $y_{true} \neq y_{pred}$ $(|y - y_{true}| \geq 0.5)$

$$\mathcal{L} = \frac{1}{1 + 1/(e^{|y_{true} - y|}) - 1}.$$

After dropping constants, we get:

$$\mathcal{L} = \frac{1}{1/|y_{true} - y|} = |y_{true} - y|. \quad (9)$$

For $y_{true} = 0$, we have the following:

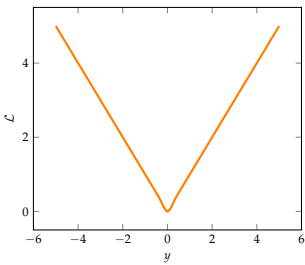

**Fig. 4.** Loss Function $\mathcal{L}$ vs. $y$ — Case 2a

For $y_{true} = 1$, we have the following:

In the first case, the plot of loss value $\mathcal{L}$ versus the predicted conditional class probability $y$ was non-convex. In the second and third cases, the plots of loss value $\mathcal{L}$ versus the predicted conditional class probability $y$ were convex. However, since our ultimate goal is to find the value of the model parameters $\theta$ that optimize $\mathcal{L}$, we will need to expand $y$ depending on the model that we are using. In other words, we will analyze the convexity of the Xtreme Margin loss function for a simple artificial neural network. In general, this could change if we switch to another model with a different relationship between its parameters and the predicted output $y$. This property significantly helps when using this loss function for deep learning, particularly when using SGD as the optimizer.

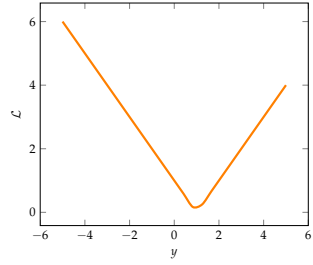

**Fig. 5.** Loss Function $\mathcal{L}$ vs. $y$ — Case 2b

## 10. CONVEXITY ANALYSIS FOR AN ANN

Consider an Artificial Neural Network (ANN) with 2 layers (1 input and 1 output layer), $n$ input features $x_1, ..., x_n$, 1 target variable, $n + 1$ nodes, no bias, and the **sigmoid** function as the output activation function.

We define the predicted probability $y$ to be the following:

$$y = sigmoid\left(\sum_{i=1}^{n} w_i x_i\right). \quad (10)$$

Substituting this value for both cases above, we get the following:

**Case 1.** $y_{true} = y_{pred}$ $(|y - y_{true}| < 0.5)$

$$\mathcal{L} = \frac{1}{1 + (2(sigmoid\left(\sum_{i=1}^{n} w_i x_i\right)) - 1)^2}. \quad (11)$$

**Case 2.** $y_{true} \neq y_{pred}$ $(|y - y_{true}| \geq 0.5)$

$$\mathcal{L} = \left| y_{true} - sigmoid\left(\sum_{i=1}^{n} w_i x_i\right) \right|. \quad (12)$$

Recall that we stated earlier that in general, the loss function may or may not be convex depending on the machine learning model chosen. However, the Xtreme Margin loss function is non-differentiable regardless of the model chosen, due to the absolute values in the denominator of the loss function $\mathcal{L}$.

## 11. CONTINUITY AND DIFFERENTIABILITY

### A. Continuity Analysis

All of the functions above are continuous and smooth. The loss function $\mathcal{L}$ can take on any value depending on the parameter values $w_1, ..., w_n$.

## B. Differentiability Analysis

When a model correctly predicts an instance's label, the function is differentiable (Figure 3). However, when it does not correctly predict an instance's label, the function may not be differentiable at all values of $y$. For instance, when $y_{true} = 1$, the $y$ value that minimizes the value of $\mathcal{L}$ in Figure 5 occurs at $0 < y < 1$. Since the loss function $\mathcal{L}$ is not differentiable at all points, gradient descent techniques cannot be performed with this loss function, as the gradient cannot be computed at some values of $y$. Both the SGD and Adam optimizers rely on gradient-based optimization methods, and hence, should not plainly be used with the Xtreme Margin loss function.

## 12. NON-CONVEX/DIFFERENTIABLE OPTIMIZATION

Now that we have shown that the Xtreme Margin loss function is non-convex and non-differentiable for the basic artificial neural network construction above, we will list out some compatible optimization techniques. Since $\mathcal{L}$ is sub-differentiable, the $\epsilon$-subgradient or the negative subgradient methods can be used, which are optimization algorithms that can be applied to any minimization problems. The following is the update rule for the negative subgradient method:

$$\theta^{t+1} \leftarrow \theta^t - \alpha * g_{\mathcal{L}}^t. \tag{13}$$

Note that the update rule above resembles the gradient descent update rule. $\theta^t$ represents the parameter value at time $t$, the $\alpha$ represents the learning rate, and $g^t$ represents any computable *subgradient* of $\mathcal{L}$ at time $t$ (not the plain gradient that gradient descent uses).

We can verify that a vector is a subgradient of an arbitrary loss function $\mathcal{L}$ at a parameter value $\theta_0$ if the following holds for all parameter values $\theta$:

$$\mathcal{L}(\theta) \geq \mathcal{L}(\theta_0) + g^T(\theta - \theta_0). \tag{14}$$

Applying this method will ultimately result in a value of $y$ that minimizes the loss function $\mathcal{L}$. The paper by Boyd et al. [2] states that the subgradient update rule is guaranteed to converge to within some reasonable range of the minimum value of the loss function $\mathcal{L}$ for a constant step size and a constant step length.

The following is the negative subgradient descent update rule algorithm pseudocode:

**Algorithm 1.** Subgradient Descent Update Rule

**Require:** $\alpha > 0$
  **for** $t = 1..T$ **do**
    Compute loss $\mathcal{L}(x)$ at time $t$
    Compute subgradient $g_t$ from $\mathcal{L}(x)$
    $x_{t+1} = x_t - \alpha * g_t$    ▷ Subgradient Descent Update Rule
    Output $x_T$

## 13. TUNING BIAS

The *bias* of a model is a systematic error during the predictions made for a particular dataset. The bias-variance decomposition states that the error can be broken down into three components: the bias of the model, the variance of the model's predictions, and random noise $\epsilon$. The bias of a model is defined as follows, where $\mathbb{E}[\hat{\theta}]$ is the expected value of the models in the population and $\theta$ is the model of interest:

$$Bias(\theta, \hat{\theta}) = \left(\mathbb{E}[\hat{\theta}] - \theta\right)^2. \tag{15}$$

$$Error(\theta) = Bias(\theta, \hat{\theta}) + Variance(\hat{\theta}) + \epsilon. \tag{16}$$

The degree of bias of a machine learning model can be tuned with the hyperparameters $\lambda_1$ and $\lambda_2$. Below, we compare the bias as the hyperparameters increase with the above neural network model trained for 50 epochs.

| Loss Function | $\lambda_1$ | $\lambda_2$ | Bias |
|---|---|---|---|
| Xtreme Margin | 1 | 50 | 0.170159 |
| Xtreme Margin | 1 | 100 | 0.193041 |
| Xtreme Margin | 100 | 1 | 0.197346 |
| Xtreme Margin | 2 | 2 | 1.058191 |
| Binary Cross-Entropy | N/A | N/A | 0.112281 |
| Hinge | N/A | N/A | 0.652224 |

As we can see above, the version of Xtreme Margin with $\lambda_1 = 1$ and $\lambda_2 = 50$ outperformed the hinge loss, but came up short to the binary cross-entropy loss. For this particular dataset, having the model learn the instances within the default class well resulted in the minimization of the model's bias.

## 14. TRAINING CONVERGENCE

Below, we compare the training and testing accuracies of different configurations of the Xtreme Margin loss functions with the binary cross-entropy loss function on the Ionosphere dataset. Each of the following plots are generated by training the neural network model above for 50 epochs.

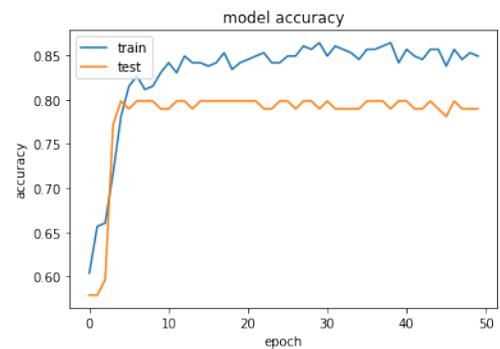

**Fig. 6.** Xtreme Margin ($\lambda_1 = 2, \lambda_2 = 2$) Evaluation Curve

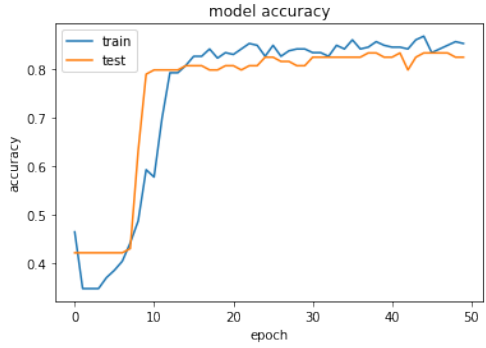

**Fig. 7.** Xtreme Margin ($\lambda_1 = 2, \lambda_2 = 100$) Evaluation Curve

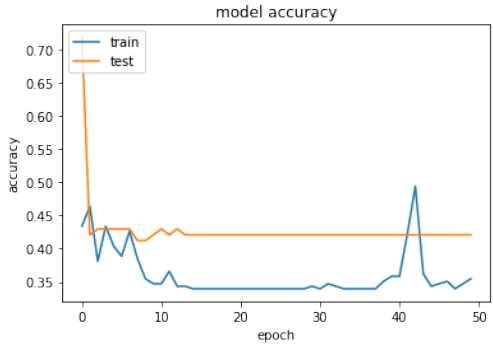

**Fig. 8.** Xtreme Margin ($\lambda_1 = 2, \lambda_2 = 500$) Evaluation Curve

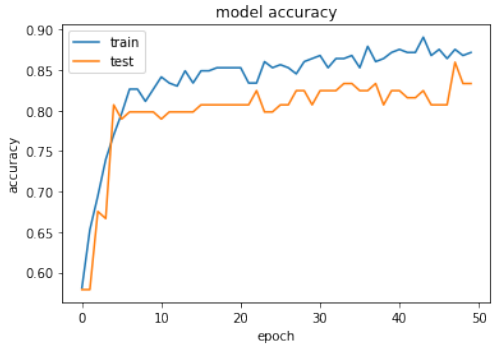

**Fig. 9.** Binary Cross-Entropy Evaluation Curve

Comparing the above plots, the model trained on the binary cross-entropy loss function achieved the highest accuracy on the Ionosphere dataset. However, it slightly overfit, as the testing accuracy was consistently below the training accuracy after 10 epochs. The Xtreme Margin loss function with $\lambda_1 = 2$ and $\lambda_2 = 100$ performed the best from a model generalization standpoint, as it did not underfit nor overfit. This is the advantage of the tunable component of the Xtreme Margin loss function — depending on the dataset, researchers and practitioners can control the degree of underfitting and overfitting.

## 15. CONDITIONAL ACCURACY

We define *conditional accuracy* as the accuracy of the data points belonging to a particular class. For instance, if a dataset $X$ contains data points belonging to 2 possible classes, and the set of data points $ND$ of them belonging to class '0' and the set of data points $D$ belonging to class '1', then the accuracy conditioned on class '0' would be as follows:

$$\text{Conditional Accuracy} = \frac{\sum_{x \in ND} \mathbb{1}_{[f(x)=0]}}{|ND|}. \quad (17)$$

On the Ionosphere dataset used for our experiment, even though the binary cross-entropy loss function achieved a higher mean cross-validation accuracy compared to the Xtreme Margin loss function, its conditional accuracy cannot be manually controlled, as it is internally chosen during the training process on the loss function. In some situations, it suffers from a low conditional accuracy for one or both classes.

The following figure shows the decision boundaries that resulted from training a shallow, less-dense neural network model with the Xtreme Margin loss functions with 2 predictor features. Below, we keep the value of $\lambda_1$ fixed, and we will increase the value of $\lambda_2$ and notice the impact it has on the decision boundary and the class assignments.

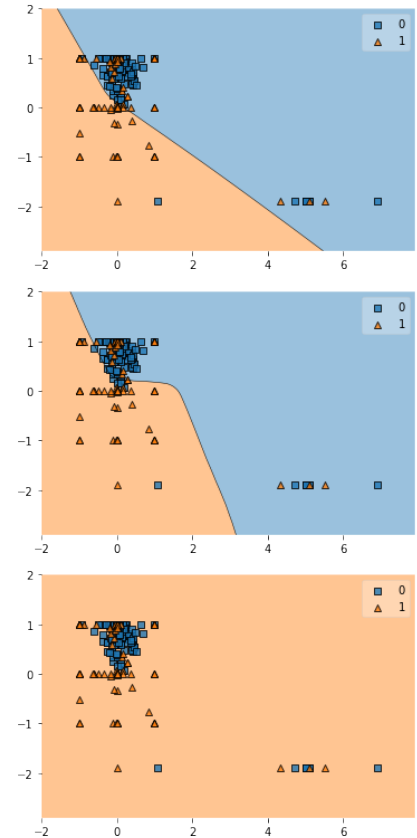

**Fig. 10.** Ionosphere Dataset Classifications with Xtreme Margin Loss (Upper: $\lambda_2 = 2$; Middle: $\lambda_2 = 100$; Lower: $\lambda_2 = 500$)

As we can see above, increasing the value of $\lambda_2$ altered the decision boundary by incentivizing the training algorithm to rack up as many orange data points within the orange region. Now, instead of holding the $\lambda_1$ hyperparameter constant, we fix the value of $\lambda_2$.

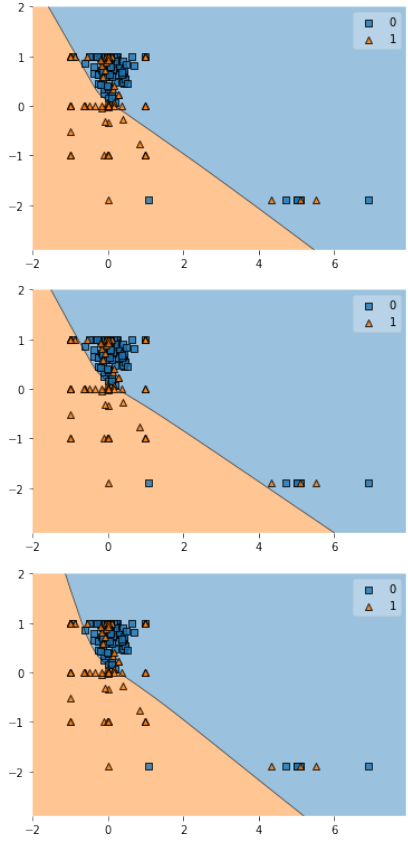

**Fig. 11.** Ionosphere Dataset Classifications with Xtreme Margin Loss (Upper: $\lambda_1 = 2$; Middle: $\lambda_1 = 100$; Lower: $\lambda_1 = 500$)

For the same configuration (i.e., number of epochs fixed), the following is the plot produced when training the same neural network model with the binary cross-entropy loss function:

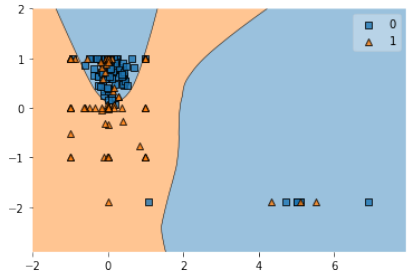

**Fig. 12.** Ionosphere Dataset Classifications with Binary Cross-Entropy Loss

## 16. THE CONDITIONAL RISK (CVAR)

The conditional risk $R(\hat{y} \mid \vec{x})$ is defined as the expectation of the loss for a particular feature vector $\vec{x}$. The conditional risk can be computed on our Xtreme Margin loss function if we were to know the confidence of the labels for each instance vector $\vec{x}$. Computing the conditional risk score may be more robust than a cost function when the true labels are not 100% certain. For example, in medical segmentation, computer vision models use training data that are labeled by radiologists in the past. However, if these labels are incorrect, or are correct with $\alpha$ probability, then this should be accounted for during a computer vision model's performance evaluation. The following is the generalized expression defined for computing the conditional risk score:

$$R(\hat{y} \mid \vec{x}; \mathcal{L}) = \sum_y \mathcal{P}(Y = y \mid X = \vec{x}) \cdot \mathcal{L}(y, \hat{y}). \quad \textbf{(18)}$$

Substituting the Xtreme Margin loss function into $\mathcal{L}(y, \hat{y})$, we derive the complete expression.

$$R(\hat{y} \mid \vec{x}; \mathcal{L}) = \sum_y \left[ \mathcal{P}(Y = y \mid X = \vec{x}) \cdot \frac{1}{1 + \sigma(y, y_{true}) + \gamma} \right]. \quad \textbf{(19)}$$

## 17. PRECISION AND RECALL METRICS

The binary cross-entropy and hinge loss functions do not have a specialized method to control the number of true positives, false positives, and false negatives of the model. This is where Xtreme Margin comes in. By controlling the ratio of $\lambda_1$ to $\lambda_2$, the precision and recall scores are able to be increased above those produced by the binary cross-entropy loss function.

| Loss Function | Lambdas | Precision | Recall |
|---|---|---|---|
| Xtreme Margin | 1; 50 | 0.9077 | 0.6667 |
| Xtreme Margin | 50; 1 | 0.9310 | 0.6000 |
| Xtreme Margin | 5; 5 | 0.9286 | 0.6778 |
| Binary Cross-Entropy | N/A | 0.9118 | 0.6556 |
| Hinge | N/A | 0.9649 | 0.6111 |

As we can see in the table above, the Xtreme Margin loss function with $\lambda_1 = 5$ and $\lambda_2 = 5$ achieved the highest recall compared to the other Xtreme Margin configurations and the binary cross-entropy and hinge loss functions. For precision, the hinge loss function achieved the best score, and following that was Xtreme Margin. The tunable component of Xtreme Margin enables practitioners to choose whether they want to maximize precision or recall. Since there is a tradeoff between precision and recall (as the precision increases, the recall decreases and vice versa), one has to place more emphasis on a particular metric depending on the use case.

For example, in healthcare applications, a patient having COVID-19 and the computer vision algorithm predicting that they do not have COVID-19 (a false negative) is more costlier than a patient not having COVID-19 and the computer vision algorithm predicting that they do have COVID-19 (a false positive). Depending on the application, the emphasis placed on precision and recall will alter, and this can be controlled by the tunable hyperparameters of the Xtreme Margin loss function based on the use case.

| Loss Function | Lambdas | AUC |
|---|---|---|
| Xtreme Margin | 1; 50 | 0.8881 |
| Xtreme Margin | 50; 1 | 0.8654 |
| Xtreme Margin | 5; 5 | 0.8857 |
| Binary Cross-Entropy | N/A | 0.8761 |
| Hinge | N/A | 0.7985 |

For the above table, we compared the AUC metrics across the 5 loss functions. For the dataset and the particular neural network model used, all the Xtreme Margin loss function variants resulted in the highest AUC scores compared to the binary cross-entropy and hinge loss functions.

As we increase the value of $\lambda_2$, the AUC score tends to increase. This is because $\lambda_2$ is directly proportional to how likely an instance belonging to the default class will be correctly predicted. The AUC metric measures the probability that a randomly chosen instance from the default class has a higher conditional probability than that for a randomly chosen instance from the non-default class. If $\lambda_2$ is higher, even the data points that were on the fence to being part of the default class will have their classifications changed from the non-default class to the default-class. This will in turn naturally increase the probability that defines the AUC metric.

## 18. APPLICATION

Consider the application of detecting the presence of a volcano from a satellite image. Suppose two computer vision classification models, model $A$ and model $B$, are trained to make predictions, and the accuracy of model $A$ is greater than that of model $B$. Even though the accuracy of model $A$ is greater than that of model $B$, it is important to consider the conditional accuracy. In this situation, a false negative is more costlier than a false positive. If the volcano erupts and residents near the site of eruption are not informed, there will likely be more injuries and deaths than if the residents were notified that the volcano would erupt, even though it did not end up erupting.

The accuracy score considers both true positives and true negatives. In this case, we should not place as much weight on the true negative instances as we should place on the true positive instances. This is where conditional accuracy maximization comes into picture. By focusing on maximizing the accuracy for the positive class, the model will confidently predict those data points, even if it is at the expense of the accuracy of the data points belonging to the negative class. And, as described earlier in this paper, we can apply conditional accuracy maximization by increasing the ratio between the tunable hyperparameters $\lambda_1$ and $\lambda_2$ of the Xtreme Margin loss.

## 19. CONCLUSION

The Xtreme Margin loss function can be used for any binary classification problems, regardless of the model you choose or any other settings. It gives practitioners additional flexibility in their training process, akin to the Support Vector Machine (SVM) regularization hyperparameter $\lambda$. This loss function works better with some models than others.

As discussed above, the implementation presented in this paper works only for a batch size of 1. This can be extended to a greater batch size by applying more complex tensor operations. The following is the complete end-to-end pipeline of machine learning training with the loss function $\mathcal{L}$:

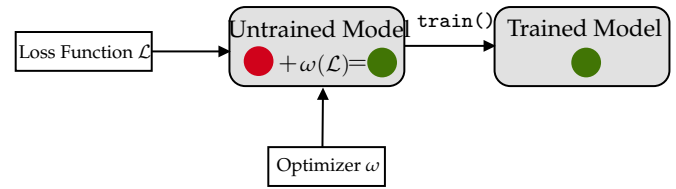